\documentclass[preprint,12pt]{elsarticle}

\usepackage{graphicx} 
\usepackage{amsmath} 
\usepackage{hyperref}
\usepackage{algorithm}
\usepackage{algpseudocode}
\usepackage{graphicx}  % in preamble
\usepackage{tabularx}
\usepackage{caption}
\usepackage{lscape} 
\usepackage{graphicx}
\usepackage{pdflscape}
\usepackage{caption}
\usepackage{xcolor}
\usepackage{float}
\usepackage{adjustbox}
\usepackage[margin=0.5in]{geometry}
\usepackage{array}
\usepackage{booktabs}
\usepackage{graphicx}
\usepackage{caption}
\usepackage{float}
\usepackage{lipsum} % optional for dummy text
\usepackage{graphicx}
\usepackage{caption}
\usepackage{array}
\usepackage{graphicx}
\usepackage{caption}
\usepackage{adjustbox}
\usepackage{geometry}
\usepackage{array}
\definecolor{darkgreen}{RGB}{0,100,0}
\usepackage{amssymb}
\usepackage{amsmath}
\newcommand{\imgcell}[1]{\includegraphics[width=0.18\linewidth, height=3cm, keepaspectratio]{images/#1}}

\newcommand{\txtcell}[1]{\begin{minipage}[t]{0.18\linewidth}\raggedright\scriptsize #1\end{minipage}}

\journal{Journal}

\begin{document}

\begin{frontmatter}

\title{Beam-Guided Knowledge Replay for Knowledge-Rich \\Image Captioning using Vision-Language Model
%Enhanced Model for Real World Knowledge Image Captioning
}
%KnowCap++: Improving Knowledge Grounding in Image Captions with Attention and Beam Search
%ReCap: Replay-Augmented Captioning with Beam Search and Visual Self-Attention

\author[1,2]{Reem AlJunaid}
\ead{reem.aljunaid.94@gmail.com}

\author[1,3]{Muzammil Behzad\corref{cor1}}
\ead{muzammil.behzad@kfupm.edu.sa}

\cortext[cor1]{Corresponding author}

\affiliation[1]{organization={King Fahd University of Petroleum and Minerals},
            country={Saudi Arabia}}

\affiliation[2]{organization={Imam Abdulrahman Bin Faisal University},
            country={Saudi Arabia}}
\affiliation[3]{organization={SDAIA-KFUPM Joint Research Center on Artificial Intelligence},
            country={Saudi Arabia}} 

%% Abstract
\begin{abstract}
Generating informative and knowledge-rich image captions remains a challenge for many existing captioning models, which often produce generic descriptions that lack specificity and contextual depth. To address this limitation, we propose KRCapVLM, a knowledge replay-based novel image captioning framework using vision-language model. We incorporate beam search decoding to generate more diverse and coherent captions. We also integrate attention-based modules into the image encoder to enhance feature representation. Finally, we employ training schedulers to improve stability and ensure smoother convergence during training. These proposals accelerate substantial gains in both caption quality and knowledge recognition. Our proposed model demonstrates clear improvements in both the accuracy of knowledge recognition and the overall quality of generated captions. It shows a stronger ability to generalize to previously unseen knowledge concepts, producing more informative and contextually relevant descriptions. These results indicate the effectiveness of our approach in enhancing the model’s capacity to generate meaningful, knowledge-grounded captions across a range of scenarios.
\end{abstract}

%% Keywords

\end{frontmatter}

\section{Introduction} 
Image captioning is the task of generating descriptions of images using computer vision and natural language processing techniques~\cite{anderson2018bottom, cornia2020meshed, vinyals2015show}. It has a variety of applications, such as enhancing the content understanding of multimedia~\cite{qu2024visually, liu2024image} and helping visually impaired people ~\cite{stefanini2022show, yu2023quality, ahsan2021multi, dognin2020image, gurari2020captioning, faurina2023image}. Current existing models, however, often produce generic captions which lack real-world concepts like contextual details and named entities ~\cite{jia2023image, nikiforova2020geo, zhao2019informative, vo2022nocrek, zhao2021boosting, lu2018entity, li2023evcap, chen2023retrieval, zhang2023fashion, ayesha2022generating}. This missing knowledge often represents key information in understanding the content of the image. Furthermore, this detailed knowledge can also enhance the performance of other models that rely on the output of image captioning systems, such as question answering systems ~\cite{wu2018joint, wu2019generating, salaberria2023image, ozdemir2024enhancing, wu2016image}.\\

Several efforts have been placed in order to enhance the descriptions with real-world knowledge~\cite{nikiforova2020geo, tran2016rich, whitehead2018incorporating}. However, most of these works are limited by using external resources, such as image metadata or object recognition models, to detect existing entities before generating the descriptions. To overcome that, the Vision-Language Pretrained (VLM) models ~\cite{li2020unicodervl, radford2021clip} present a powerful solution for this era. They were trained on massive data and can capture diverse real-world knowledge \cite{li2020unicodervl, li2022blip, radford2021clip, wang2022ofa, wang2021simvlm, zhang2021vinvl}. Despite their tremendous performance, VLM models suffer from two main problems: (1) zero-shot inference leading to safe but low-quality descriptions \cite{cheng2023knowledge, dai2022object, zhao2025mitigating}, and (2) knowledge hallucination due to the noise in image-text pairs in pre-training \cite{cheng2023knowledge, dai2022object}. Additionally, fine-tuning VLM models on downstream tasks introduces a ``generic bias'' that restricts their expression of detailed knowledge \cite{zhao2025mitigating}.\\

Addressing these limitations, the Knowledge guided Replay (K-Replay) framework is proposed in \cite{cheng2023beyond}. K-Replay preserves the original model structure while helping VLM models retain knowledge during fine-tuning on downstream tasks. It is done by selecting knowledge-rich samples from the pretraining data and computing a coverage loss based on the existence of required keywords in the sentence to reinforce memory of this knowledge, and hence, avoid generic descriptions. Moreover, to reduce hallucination and ensure faithful descriptions, a knowledge distillation-based constraint is applied from a fine-tuned model. With strong performance, especially in unseen scenarios, K-Replay effectively helps the model to recall and express pre-learned knowledge.\\ 

Despite these strengths, K-Replay has several limitations. First, it generates pseudo-captions for replay samples using greedy decoding, which tends to produce low-diversity and overly generic sentences. These captions may fail to capture detailed knowledge, thereby weakening the replay mechanism's effectiveness. Second, the combination of multiple loss terms in its framework often leads to unstable training dynamics and convergence issues. Finally, K-Replay lacks explicit self-attention among image patches, limiting its ability to capture fine-grained visual patterns essential for real-world knowledge.\\

Furthermore, this work focuses on enhancing VLM models to produce knowledge-rich and accurate image captions by evaluating using the KnowCap \cite{cheng2023beyond} dataset. As such, the scope of knowledge assessed is limited to the types of real-world knowledge represented by the predefined keyword categories in the KnowCap dataset. These include four categories, foods, brands, landmarks, and movie characters.  The study does not attempt to capture or evaluate all forms of general knowledge but it focuses on the model’s ability to incorporate specific knowledge keywords, as defined by this benchmark, into the generated captions.

%In this work, we enhance the K-Replay framework \cite{cheng2023beyond} by replacing greedy decoding with beam search to generate higher-quality pseudo-captions during replay. Second, we introduce learning rate schedulers to stabilize training and promote smoother convergence. Finally, we integrate attention layers into the model architecture to improve its ability to focus on relevant image regions and contextual cues. These enhancements help improve the model’s capacity to produce detailed and accurate knowledge-based captions. 

\section{Literature Review} 
Image captioning has evolved through various techniques, starting from traditional encoder-decoder architecture to transformer-based architecture~\cite{yang2020survey, zhang2023survey, li2021image, chao2022survey}. In traditional techniques, images are encoded with convolutional neural networks (CNNs) and texts decoded with recurrent neural networks (RNNs), converting visual features into language linearly. Even though these models are effective, they often fail to capture extended dependencies within captions and the complexity of visual scenes. On the other hand, transformer-based frameworks offer attention mechanisms for identifying semantic relations within visual scenes rather than considering them separately. Using self-attention, transformers create accurate, knowledge-rich captions that capture detailed scenes despite lengthy or complex descriptions. Compared to traditional approaches, these transformer-based models provide superior accuracy and knowledge-rich captions.

Traditional encoder-decoder architecture plays a key role in the development of image captioning models. For instance, Vinyals et al. ~\cite{vinyals2015show} pioneered the use of a deep learning encoder-decoder architecture for image captioning. Their approach involved using pre-trained CNNs for image encoding, with the last hidden layer output passed into an RNN to generate descriptive captions. Their work provided a baseline for subsequent studies, which have expanded and refined the model by incorporating various enhancements, including the visual attention mechanism introduced by ~\cite{xu2015show}. Another popular framework that serves as a baseline for many image captioning models is the Bottom-Up and Top-Down framework~\cite{anderson2018bottom}. By leveraging Faster R-CNN, it combines two types of attention: Bottom-Up, which identifies important parts of the image using feature vectors, while Top-Down decides which parts to focus on while generating the next word of the caption. Similarly, Huang et al.~\cite{huang2019attention} developed the attention-on-attention (AoA) framework, which improves traditional attention by adding another layer to help the model choose more relevant words based on context. Another encoder-decoder-based framework is the Deep Hierarchical Encoder-Decoder Network (DHEDN)\cite{xiao2019deep}. DHEDN comprises a three-layer LSTM architecture: S-LSTM handles text encoding, VSE-LSTM combines visual and textual features into a common semantic space, and SF-LSTM is responsible for generating captions.

Most recently, research in image captioning has shifted towards transformer-based architecture. The self-attention in deep learning models enables simultaneous processing of images and captions, thus helping generate more accurate and knowledge-rich descriptions ~\cite{zhang2023survey}. Unlike RNNs, transformers process sequences in parallel, making training much faster and more scalable. In this regard, Vaswani et al.~\cite{jia2015guiding} introduced self-attention for efficient parallel processing, replacing recurrence entirely. The model treats image features as input tokens and uses standard encoder-decoder layers for generation. However, the model lacks visual enhancements. In a follow up work, Cornia et al. presented the M2 meshed-memory transformer, which introduced memory vectors to encode multi-level relationships among image regions. The encoder analyzes the image regions and their connections, while the decoder generates captions dynamically.

Building on advances in natural language processing (NLP), the BERT architecture ~\cite{devlin2019bert} employs a masked language modeling objective to enable attention in both directions. This design choice has substantially boosted its ability to understand context. Although BERT was originally developed to perform text-based tasks, its architecture has inspired its use in multimodal applications such as image captioning. Several researchers have developed hybrid models combining BERT-like components with visual encoders to link visual and textual features ~\cite{chen2017structcap,chen2017groupcap}. These models use pre-trained language representations to enhance the connection between images and captions.

Recent advancements in image captioning have been driven by vision-language models (VLMs) models. These models are initially trained on a large-scale dataset using self-supervised learning and later adapted to perform a downstream task. One of the most utilized pre-trained models is Contrastive Language-Image Pre-Training (CLIP) by Radford et al.~\cite{radford2021learning}. CLIP has been trained using a contrastive loss on a vast dataset of image-caption pairs. Building upon CLIP, Mokady et al.~\cite{mokady2021clipcap} proposed ClipCap, a method that maps CLIP image embeddings into a prefix used to condition a pre-trained language model (GPT-2) for caption generation. The method demonstrated the effectiveness of combining visual and linguistic pre-trained models. Another similar VLM model is OSCAR (Object-Semantics Aligned Pretraining for Vision-and-Language Tasks) introduced by Li et al.~\cite{li2020oscar}. OSCAR improves cross-modal representation learning by injecting object tags as anchor points during pretraining. This approach enhances the alignment between image regions and descriptions, thus enhancing the image captioning performance. Recently, several VLMs, such as BLIP \cite{li2022blip}, GIT \cite{wang2022git}, and OFA \cite{wang2022ofa}, have been trained for image captioning tasks and have outperformed existing methods across multiple benchmarks.

However, many studies have shown that the generated captions remain generic and lack real-world knowledge. To overcome this limitation, several studies have attempted to enrich captions with knowledge through a retrieve-and-generate framework using external resources. For example, entity-aware models integrate named entities into the captioning pipeline, either through predefined templates \cite{lu2018entity} or by modifying decoder architectures \cite{biten2019good}, \cite{tran2020transform}. Other methods leverage visual recognition outputs \cite{tran2016rich}, \cite{zhao2019informative} or contextual metadata such as geo-location \cite{nikiforova2020geo} to guide caption generation. However, these methods require a lot of extensive supervision and annotated datasets. Other studies fine-tuned VLMs for image captioning. However, this approach, on the contrary, suffers from generic bias, which limits its informativeness.

\subsection{Challenges and Motivations} 
Despite advances in image captioning, existing models still struggle to generate knowledge-rich and context-aware descriptions. Transformer and VLM-based models often produce generic captions due to fine-tuning on limited datasets, which can override pre-trained knowledge. Additionally, many methods rely on external tools to inject knowledge, making them less scalable. Hallucination and forgetting of real-world knowledge remain key challenges, especially in zero-shot scenarios, highlighting the need for better knowledge retention and expression during fine-tuning. In contrast, this paper leverages a VLM model to mitigate generic bias and hallucinations, while enhancing informativeness.

The baseline model used in this work is the K-Replay framework \cite{cheng2023beyond}, which was originally proposed to address the issue of knowledge retention during the fine-tuning of VLMs for image captioning. While this framework showed promise in preserving real-world knowledge during training, it has notable limitations. One major issue lies in its method of generating pseudo-captions for the replay samples. The original implementation uses greedy decoding, which often leads to low-diversity and generic outputs. These captions may lack detail and fail to fully represent the knowledge concepts intended for replay, thereby limiting the effectiveness of the knowledge-guided learning process.
Another challenge observed with the baseline model is the instability of the training process. The K-Replay framework combines multiple loss terms, a cross-entropy loss, a replay loss, and a distillation loss, which can lead to fluctuating training curves and difficulty in convergence. This lack of smooth optimization raises the need for mechanisms that can balance learning dynamics. Additionally, while the original OFA architecture includes attention mechanisms across both visual and textual modalities, it does not explicitly model self-attention among image patches prior to fusion with text. This may limit the model’s ability to capture detailed visual patterns that are important for understanding real-world knowledge. These limitations highlight the need for further refinement.

\subsection{Contributions} 
This research extends the KnowCap framework by incorporating beam search decoding, attention-based mechanisms, and training schedulers, resulting in improved caption quality and greater accuracy in knowledge recognition within images. The findings demonstrate significant improvements in model performance, especially in its ability to generalize to unseen knowledge categories. These advancements have the potential to make a significant impact in a wide range of areas including robotics, autonomous systems, medical imaging and diagnostics, and assistive technologies, where precise and context-aware image descriptions are essential.
Consequently, the primary objective of this research is to enhance image captioning models by enabling them to generate more accurate and knowledge-rich descriptions. To achieve this, the salient features of our work are as follows:
\begin{itemize}
    \item We address the challenge of knowledge hallucination in fine-tuned VLM models by developing a more effective fine-tuning strategy.
    \item We extend the K-Replay framework by integrating beam search to generate higher-quality pseudo-captions, thereby enriching the training data.
    \item We introduce additional attention layers to strengthen the model’s ability to focus on semantically important regions in the image and contextual information.
    \item We incorporate learning rate schedulers to improve training stability and ensure better convergence during model optimization.
    \item We evaluate the proposed improvements through extensive experiments on the KnowCap and COCO datasets to validate their effectiveness.
\end{itemize}

\begin{figure*}[t]
    \centering
    \includegraphics[width=\textwidth]{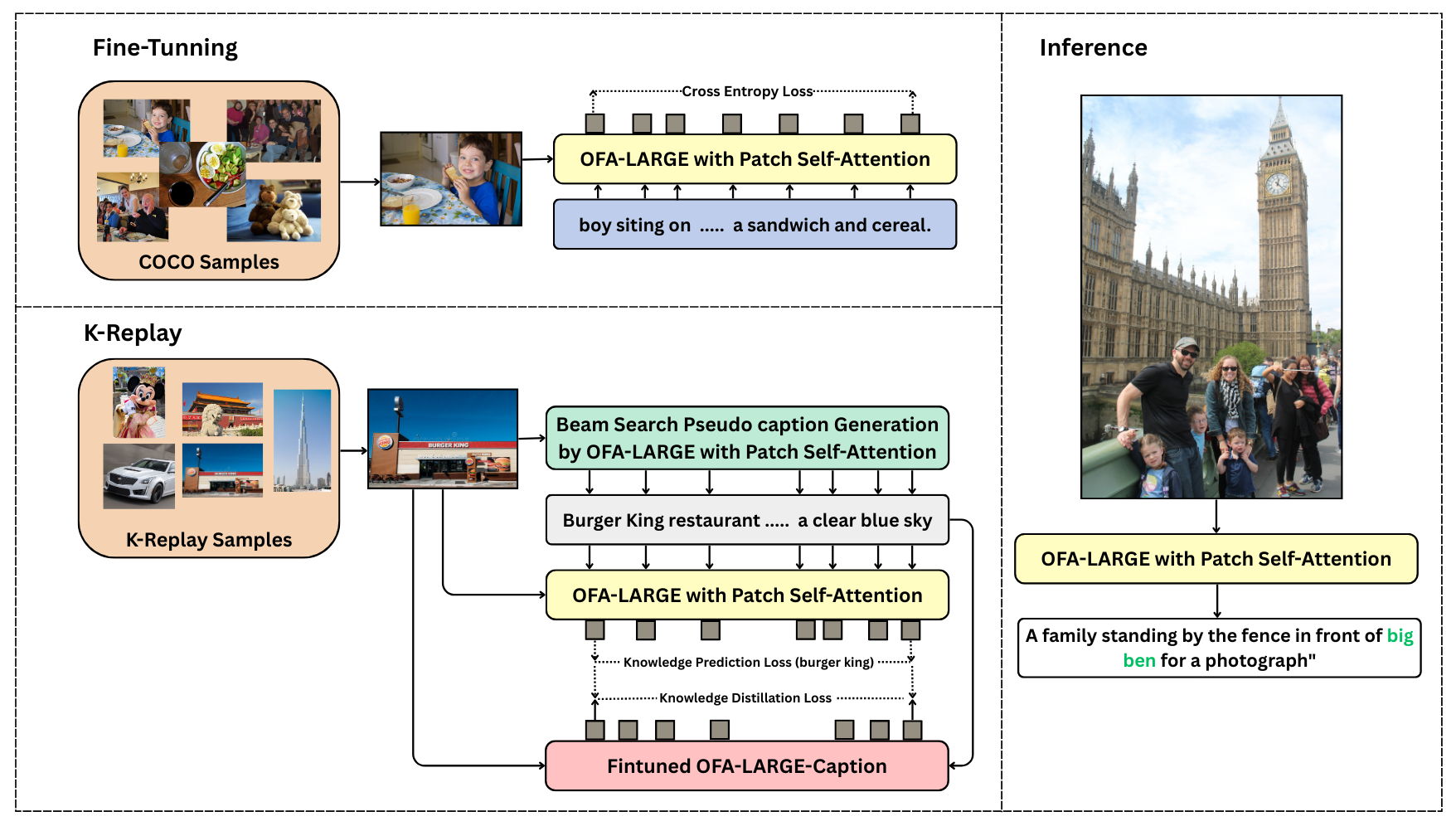}
    \caption{Illustration of our KRCapVLM framework. Knowledge replay is performed in parallel with downstream task fine-tuning to reinforce the model’s knowledge of concepts.}
    \label{fig:framework}
\end{figure*}

\section{Proposed Methodology} 
This section describes our proposed contributions for developing a knowledge-rich captioning model. It outlines our key contributions and provides an overview of our proposed knowledge-replay based captioning model using vision-language model (KRCapVLM).

\subsection{KRCapVLM Framework} 
\begin{figure}[t] % 't' asks LaTeX to place it at the top of the column
    \centering
    \includegraphics[width=11cm]{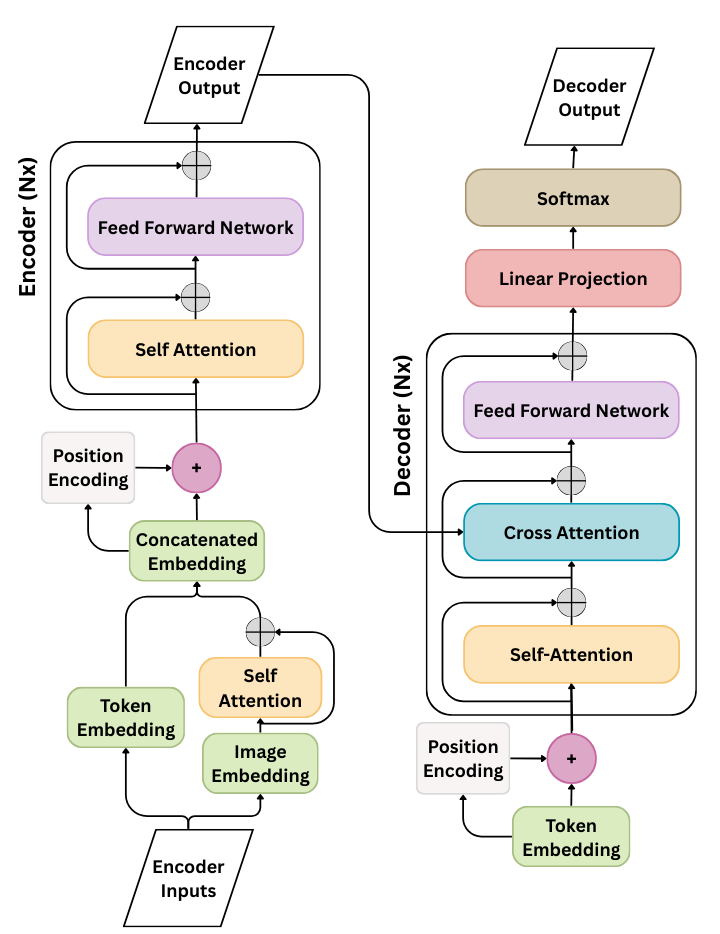}
    \caption{Custom OFA-Large model with Patch Self-Attention for Image Captioning.}
    \label{fig:model-arch}
\end{figure}

In our proposed KRCapVLM framework, we introduce three key improvements aimed at boosting the quality of pseudo-captions, improving training stability, and helping the model better understand visual content. Figure \ref{fig:framework} shows the general overview of our proposed framework. First, we propose the beam search strategy used for generating pseudo-captions. This allows the model to explore multiple caption possibilities and choose the most informative one, resulting in higher-quality replay samples during training.
Second, we enhance the encoder by adding a self-attention layer over the image patch embeddings before they are fused with text as illustrated in Figure \ref{fig:model-arch} showing the architecture of OFA-Large model with patch self-attention. In the original framework, the encoder processed image patches without explicitly modeling the relationships between different regions of the image. By incorporating self-attention, the model can attend to relevant areas of the image based on their semantic importance, helping it focus on detailed visual patterns. 
Lastly, we apply learning rate schedulers, specifically using cosine annealing, to control the pace of learning. Since the training involves multiple loss components, we noticed that the loss curves were previously unstable. By dynamically adjusting the learning dynamically over time, the training process becomes more stable and converges more smoothly. This prevents drastic updates and helps the model avoid local minima. This contributes further to better final performance and more efficient training. These proposals collectively can help generate accurate, and knowledge-rich captions. 

\subsection{Algorithm and Implementation} 
KRCapVLM utilizes a continual learning strategy designed to prevent catastrophic forgetting. It works by selectively replaying knowledge from previous tasks while learning new ones. This helps maintains a buffer of knowledge-rich examples from earlier tasks and helps reuse them during training on new tasks to ensure the model retains essential prior knowledge. In our setup, we select a pretrained VLM, OFA-Large \cite{ofa_huggingface}, which was pretrained on multi-modal tasks and datasets, including CC12M \cite{changpinyo2021conceptual}. The model we aim to train is denoted as \(M_\theta\), while a fine-tuned version, OFA-Large for image captioning, is used as a teacher model, denoted \(M_{ref}\). We use CC12M and COCO datasets to train the model\(M_\theta\). The COCO samples contain image-caption pairs \(\mathcal{S}_c\), and the CC12M samples contain image-keyword pairs \(\mathcal{S}_k\), which serve as replay samples.

During each training iteration, the algorithm operates on a mini-batch containing samples from both the current task (\(\mathcal{S}_c\)) and the replay buffer containing past knowledge-related samples (\(\mathcal{S}_k\)). If the current batch contains an image from \(\mathcal{S}_c\), it is passed through \(M_\theta\) to generate a caption. The generated caption is then compared with the ground truth caption, and the cross-entropy loss \(\mathcal{L}_{txt}\) is computed by comparing the logits of the predicted caption with the true caption. However, if the current batch contains an image from \(\mathcal{S}_k\), the teacher model \(M_{ref}\) is first used to generate a pseudo-caption using beam search. The image and the pseudo-caption are then fed into both \(M_{ref}\) and \(M_\theta\) to compute their logits. These logits are used to calculate the knowledge distillation loss \(\mathcal{L}_{distill}\). Additionally, the presence of the original keyword in the generated caption is cross-referenced to calculate the knowledge prediction loss \(\mathcal{L}_{kpred}\). The overall loss is the total of the cross-entropy loss \(\mathcal{L}_{txt}\), the knowledge distillation loss \(\mathcal{L}_{distill}\), and the knowledge prediction loss \(\mathcal{L}_{kpred}\). The hyperparameters \(\lambda_{k}\) and \(\lambda_{d}\) are employed to control the balance between the different losses. The step-by-step procedure of this training loop is outlined in Algorithm \ref{algo:kr_capvlm}, which provides a pseudocode representation of the K-Replay training mechanism used in our setup.

\subsection{Loss Function and Optimization} 
The proposed framework combines multiple objectives into a unified multitask learning setup. The total loss function comprises three key components as outlined below. \\

\subsubsection{Cross-Entropy Loss} 
The cross-entropy loss is used during the fine-tuning of the \(M_\theta\) model for the image captioning task using images from the COCO dataset. It measures the difference between the generated caption and the ground truth captions, with the goal of maximizing the likelihood of generating the correct caption \(t\) given an input image \(i\) under the current model parameters \(\theta\).
\begin{equation}
\mathcal{L}_{ce} = -\frac{1}{T} \sum_{t=1}^{T} \log p(w_t \mid w_{<t}, i; M_\theta).
\end{equation}

\begin{figure}[t]  % use figure to allow placement control
\centering
\resizebox{\linewidth}{!}{  % scale to 90% of text width, auto height
\begin{minipage}{\textwidth}
\begin{algorithm}[H]
\caption{Training KRCapVLM Model for Knowledge-Rich Image Captioning}
\label{algo:kr_capvlm}
\textbf{Input:} Training mini-batch $b \in \mathcal{S}_c \cup \mathcal{S}_k$, \\
\hspace*{1em} $b = \{(i_u, t_u)\}_u \cup \{(i_v, k_v)\}_v$; VLM  model with extra attention layer $M_\theta$; fine-tuned VLM  model $M_{ref}$; loss weights $\lambda_{k}$, $\lambda_{d}$ \\
\textbf{Output:} Updated model parameters $\theta'$
\begin{algorithmic}[1]
\For{each sample $[i, t, (k)]$ in $b$}
    \If{$i \in \mathcal{S}_c$}
        \State apply self-attention on image patches in $M_\theta$
        \State $z \gets M_\theta(i, t)$
        \State $\mathcal{L}_{ce} \gets \text{CE}(z, t)$
    \EndIf
    \If{$i \in \mathcal{S}_k$}
        \State $\hat{t} \gets \text{BeamDecode}(M_\theta(i), b{=}5)$
        \State $\tilde{z} \gets M_{ref}(i, \hat{t})$
        \State $z \gets M_\theta(i, \hat{t})$
        \State $\mathcal{L}_{kpred} \gets \text{KPL}(z, k)$
        \State $\mathcal{L}_{distill} \gets \text{KD}(z, \tilde{z})$
    \EndIf
\EndFor
\State $\mathcal{L}_{total} \gets \mathcal{L}_{ce} + \lambda_{k} \cdot \mathcal{L}_{kpred} + \lambda_{d} \cdot \mathcal{L}_{distill}$
\State update $\theta$ using AdamW optimizer and learning rate scheduler
\end{algorithmic}
\end{algorithm}
\end{minipage}
}
\end{figure}
\subsubsection{Knowledge Prediction Loss}
The knowledge prediction loss is introduced to encourage the model to incorporate knowledge-related tokens during caption generation. It is computed over the generated pseudo-captions for images from the CC12M dataset during caption generation by prompting the \(M_\theta\) model to identify named entities present in the image. Given an image-keyword pair \((i, k)\), where the keyword \(k\) is tokenized into BPE subwords \(\{w_1^k, w_2^k, ..., w_N^k\}\), the objective is to ensure that these tokens are reflected in the generated caption \(\hat{t} = M_\theta(i)\).
\begin{equation}
\mathcal{L}_{cov} = -\sum_{i=1}^{N} \log \sigma(p(w_i^{k})).
\end{equation}
To further prevent the model from excessively repeating the keywords, a repetition penalty is added to the loss as:
\begin{equation}
\mathcal{L}_{rep} = \sum_{i=1}^{N} \left[1 - p(w_i^{k}) \right]^2.
\end{equation}
The final knowledge prediction loss combines both components, knowledge coverage and repetition penalty.
\begin{equation}
\mathcal{L}_{kpred} = \mathcal{L}_{cov} + \mathcal{L}_{rep}.
\end{equation}

\subsubsection{Knowledge Distillation Loss}
The knowledge distillation loss helps the updated model keep the pre-trained network’s knowledge and generalization abilities as it adapts to captioning. This loss applies the Kullback–Leibler (KL) divergence ~\cite{kullback1951information} between the output logits of the teacher model (pre-trained \(M_{ref}\)) and the student model (the \(M_\theta\) model undergoing fine-tuning). It quantifies how much the student's output distribution deviates from that of the teacher. It acts as a regularization term to ensure the fine-tuning does not lead to catastrophic forgetting of the original model's learned representations.
\begin{equation}
\mathcal{L}_{distill} = D_{kl}[\varphi(z_t), \varphi(z_s)], \\
\textstyle \varphi(z_i) = \frac{\exp(z_i / T)}{\sum_j \exp(z_j / T)}.
\end{equation}

\section{Experimental Design and Evaluation} 
This section provides an overview of the datasets used, the evaluation metrics applied, and the training setup employed in our experiments. We present the results of our model and compare its performance against baseline methods to highlight its effectiveness. Additionally, an ablation study is conducted to investigate the contribution of individual components within the model, helping to better understand their impact on overall performance.
\subsection{Datasets and Preprocessing} 
convert this to para as follows

We use three datasets to evaluate our proposed approach. The first is MS-COCO \cite{chen2015microsoft}, a widely used image captioning dataset that follows the Karpathy split, comprising 113,287 training images, 5,000 validation images, and 5,000 test images, each annotated with five human-written captions. The second is the Replay CC12M Subset \cite{changpinyo2021conceptual}, a curated collection of over 20,000 image-text pairs extracted from the CC12M dataset by filtering for mentions of 122 predefined keywords. These samples are employed as replay exemplars during training to enhance learning. The third dataset is KnowCap \cite{cheng2023beyond}, which supports knowledge-enhanced captioning and includes 1,424 image-caption pairs spanning 240 knowledge categories. From these datasets, 424 samples are allocated for validation and 1,000 for testing. Within the test set, a specific subset of 520 samples is referred to as the unseen set which covers 120 knowledge categories that are not present in the predefined keyword list. This set is used to assess the model’s ability to generalize to novel knowledge concepts.

As part of preprocessing, all datasets are first formatted into the standard pycocoevalcap format. For training, we randomly selected 27,000 image-caption pairs from the COCO training set and 5,000 image-keyword pairs from the Replay CC12M subset. These were then combined and shuffled to form the 32,000-sample training set. The models are evaluated during training using both the COCO and KnowCap validation sets. For testing, we use the COCO test set, the full KnowCap test set, and the unseen subset from KnowCap.

\subsection{Performance Metrics} 
The evaluation of the generated captions was carried out using BLEU (B), METEOR~(M), ROUGE (R), CIDEr (C), and Knowledge Recognition Accuracy (Rec). BLEU ~\cite{papineni2002bleu} measures the precision of n-grams in the generated caption that appear in the ground truth captions. METEOR ~\cite{banerjee2005meteor} is similar to BLEU, but it incorporates semantic similarity by considering stemming, synonyms, and exact matches rather than just exact word matches. ROUGE ~\cite{lin2004rouge} primarily evaluates recall by comparing the overlap of n-grams, longest common subsequences (LCS), or word sequences between the generated and ground truth captions. CIDEr~\cite{vedantam2015cider} scores the generated captions by comparing the TF-IDF weighted n-gram similarity with a set of human-generated reference captions. Knowledge recognition accuracy is used to evaluate the inclusion of real-world concepts in generated captions. It measures the percentage of captions that correctly include valid knowledge-related keywords from the KnowCap dataset.

\subsection{Experiment Setup} 
All experiments were conducted using an NVIDIA A100 GPU. We used the official checkpoint of the PyTorch OFA-Large model with a batch size of 8, trained for 10 epochs using a learning rate of 7e-6 and label smoothing of 0.1. Knowledge distillation was performed using a temperature $T=16$. We report results for the best-performing model checkpoint based on validation performance on the KnowCap dataset.

\section{Results and Discussion} 

\subsection{Comparative Analysis} 
We evaluate our model on the KnowCap and MSCOCO datasets, placing primary emphasis on the CIDEr score and recognition accuracy, which best capture the quality and real-world grounding of generated captions. The results are presented in Table~\ref{table:dataset_comparison} where B1–B4, M, R, C, and Rec columns represent BLEU-1 to BLEU-4, METEOR, ROUGE-L, CIDEr, and recognition accuracy, respectively. As it is shown in Table~\ref{table:dataset_comparison} with the zero-shot setting, the base OFA model performs poorly on both datasets, achieving a CIDEr of 39.2 on KnowCap and only 22.1 on the MSCOCO dataset. It can be noticed that its recognition accuracy is also limited, with just 39.8\% on the KnowCap dataset. These results highlight the model’s inability to generalize to real-world concept captioning without fine-tuning.

\begin{table}[t!] % Use regular table environment in single-column
\centering
\caption{Comparison of different model variants on the KnowCap and COCO datasets.}
\resizebox{\linewidth}{!}{% Adjust 0.9 to 0.85 or lower for more reduction
\begin{tabular}{l|ccccccc|c|ccccccc}
\toprule
\textbf{Dataset} & \multicolumn{8}{c|}{\textbf{KnowCap\cite{cheng2023beyond}}} & \multicolumn{7}{c}{\textbf{COCO\cite{chen2015microsoft}}} \\
\midrule
\textbf{Model} & \textbf{B1} & \textbf{B2} & \textbf{B3} & \textbf{B4} & \textbf{M} & \textbf{R} & \textbf{C} & \textbf{Rec} & \textbf{B1} & \textbf{B2} & \textbf{B3} & \textbf{B4} & \textbf{M} & \textbf{R} & \textbf{C} \\
\midrule
OFA zero-shot & \textcolor{red}{32.8} & \textcolor{red}{20.0} & \textcolor{red}{13.5} & \textcolor{red}{9.4} & \textcolor{red}{11.5} & \textcolor{red}{24.6} & \textcolor{red}{39.2} & 39.80\% & \textcolor{red}{24.5} & \textcolor{red}{14.6} & \textcolor{red}{9.3} & \textcolor{red}{6.1} & \textcolor{red}{10.3} & \textcolor{red}{21.0} & \textcolor{red}{22.1} \\
OFA-Finetuned & 35.6 & 22.0 & 15.2 & 10.8 & 12.1 & 26.6 & 41.7 & \textcolor{red}{38.50\%} & \textcolor{blue}{79.6} & \textcolor{blue}{64.7} & \textcolor{blue}{50.9} & \textcolor{blue}{39.8} & 30.4 & \textcolor{blue}{59.7} & \textcolor{blue}{134.6} \\
\midrule
+K-Replay & \textcolor{blue}{58.8} & 41.5 & 30.5 & 22.7 & 20.2 & 43.0 & 90.3 & 50.40\% & 77.3 & 62.0 & 48.6 & 37.8 & 30.6 & 59.2 & 130.6 \\
+Scheduler & 58.6 & \textcolor{blue}{42.3} & \textcolor{blue}{31.7} & \textcolor{blue}{23.9} & \textcolor{blue}{20.9} & \textcolor{blue}{44.1} & \textcolor{blue}{92.6} & 54.20\% & 77.8 & 62.4 & 48.7 & 37.7 & 30.4 & 59.1 & 130.4 \\
+Beam & 57.7 & 40.7 & 29.6 & 21.9 & 20.4 & 42.5 & \textcolor{blue}{92.6} & \textcolor{blue}{63.30\%} & 77.2 & 62.1 & 48.7 & 37.9 & \textcolor{blue}{30.7} & 59.2 & \textcolor{blue}{130.8} \\
+Attention & 58.1 & 41.1 & 30.1 & 22.1 & 20.5 & 43.2 & 92.0 & 58.90\% & 77.0 & 61.7 & 48.3 & 37.6 & 30.6 & 59.0 & 129.4 \\
\bottomrule
\end{tabular}
}
\label{table:dataset_comparison}
\end{table}

Once fine-tuned (+OFA-Finetuned), the model shows substantial improvements in general captioning metrics. On the KnowCap dataset, the CIDEr score increases from 39.2 to 41.7, and a large jump can be seen from 22.1 to 134.6 on the MSCOCO dataset. However, this improvement comes with a trade-off as recognition accuracy on the KnowCap dataset slightly decreases from 39.8\% to 38.5\%, even though the overall fluency and structure of the captions improve. This indicates signs of catastrophic forgetting, where the model becomes less effective at grounding captions in real-world concepts due to overfitting to MSCOCO-style language patterns during fine-tuning. It highlights the importance of knowledge replay to retain domain-specific knowledge without compromising language generation quality.+
\begin{table}[t!]
\centering
\caption{Performance on unseen categories of the KnowCap dataset.}
\label{table:unseen_comparison}
\setlength{\tabcolsep}{6pt} % Slightly more space between columns
\begin{tabular}{lcccccccc}
\toprule
\textbf{Model} & \textbf{B1} & \textbf{B2} & \textbf{B3} & \textbf{B4} & \textbf{M} & \textbf{R} & \textbf{C} & \textbf{Rec} \\
\midrule
OFA zero-shot     & \textcolor{red}{31.7} & \textcolor{red}{19.5} & \textcolor{red}{13.5} & \textcolor{red}{9.8}  & \textcolor{red}{11.3} & \textcolor{red}{24.1} & \textcolor{red}{36.9} & 39.20\% \\
OFA-Finetuned     & 35.0 & 22.2 & 15.7 & 11.5 & 12.3 & 26.5 & 42.0 & \textcolor{red}{39.00\%} \\
\midrule
+K-Replay         & 57.9 & 40.8 & 30.5 & 23.2 & 19.6 & 42.6 & 83.9 & 45.80\% \\
+Scheduler        & \textcolor{blue}{58.7} & \textcolor{blue}{42.4} & \textcolor{blue}{31.9} & \textcolor{blue}{24.3} & \textcolor{blue}{20.8} & \textcolor{blue}{44.2} & \textcolor{blue}{91.0} & 55.60\% \\
+Beam             & 56.2 & 39.5 & 28.9 & 21.7 & 19.5 & 41.8 & 82.6 & \textcolor{blue}{57.69\%} \\
+Attention        & 57.5 & 40.8 & 30.1 & 22.5 & 20.1 & 43.0 & 84.4 & 55.00\% \\
\bottomrule
\end{tabular}
\end{table}

We can see that incorporating K-Replay leads to significant improvements on both datasets. The CIDEr score on KnowCap more than doubles by reaching 90.3, while recognition accuracy also rises significantly to 50.4\%, demonstrating that the model now retains better knowledge of real-world concepts. On the MSCOCO dataset, the CIDEr score remains strong at 130.6, showing that knowledge retention does not come at the cost of general captioning performance. The use of a learning rate scheduler (+Scheduler) brings moderate improvements in CIDEr (92.6) and a noticeable increase in the recognition accuracy (54.2\%) on the KnowCap dataset. This suggests that a more stable optimization process can positively influence both caption fluency and concept recognition. Introducing beam search decoding (+Beam) significantly increases recognition accuracy to 63.3\% which is the highest recorded on the KnowCap dataset although the CIDEr score slightly drops to 92.6. This indicates a trade-off between linguistic diversity and precision. Nevertheless, it can be seen that beam search enhances the model’s ability to recover conceptually meaningful captions. Lastly, augmenting the model with additional attention layers (+Attention) offers a balanced improvement, achieving 92.0 CIDEr and 58.9\% recognition accuracy on the KnowCap dataset. 

\begin{figure*}[p]
\centering
% Row 1
\imgcell{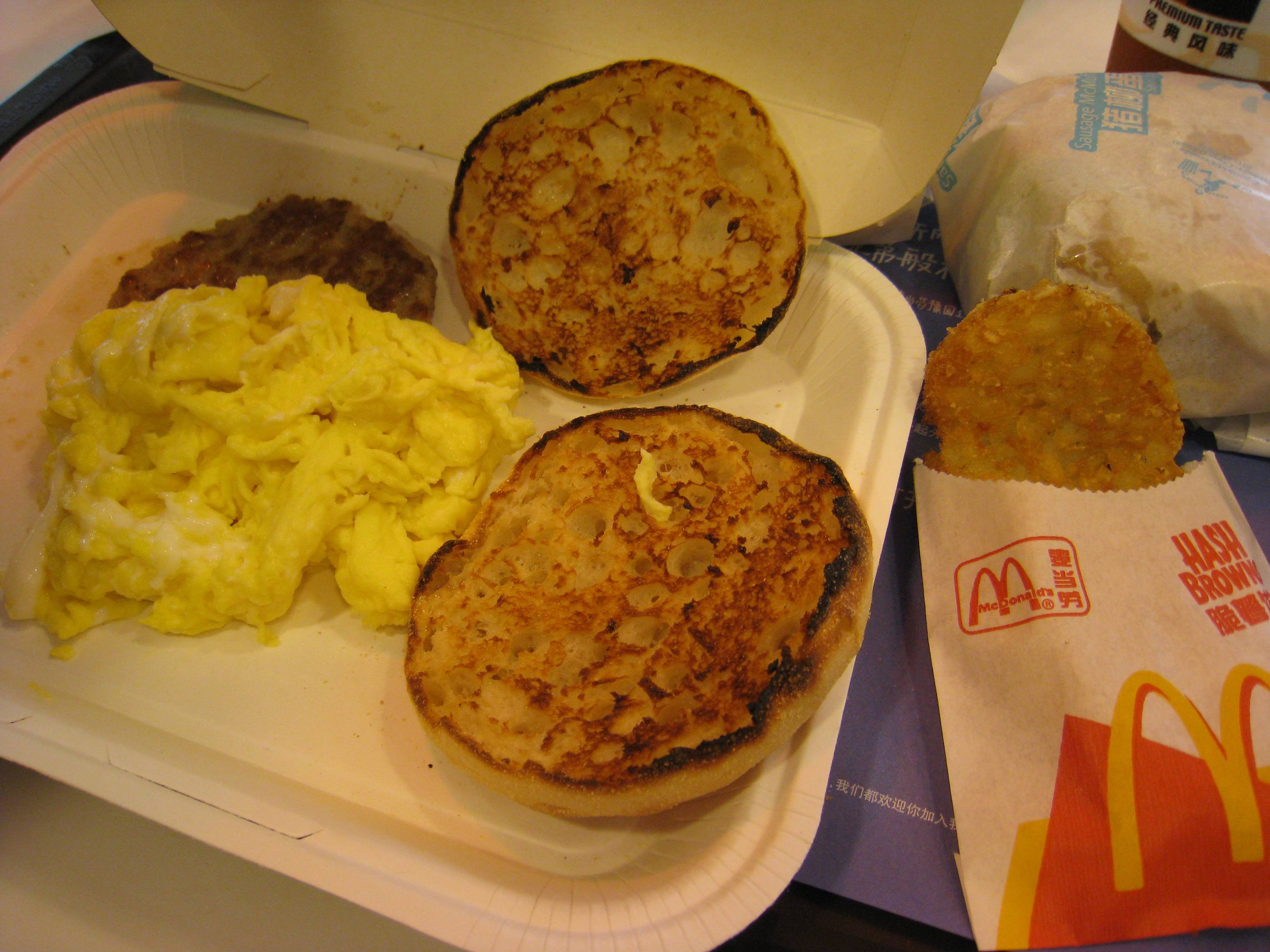}
\imgcell{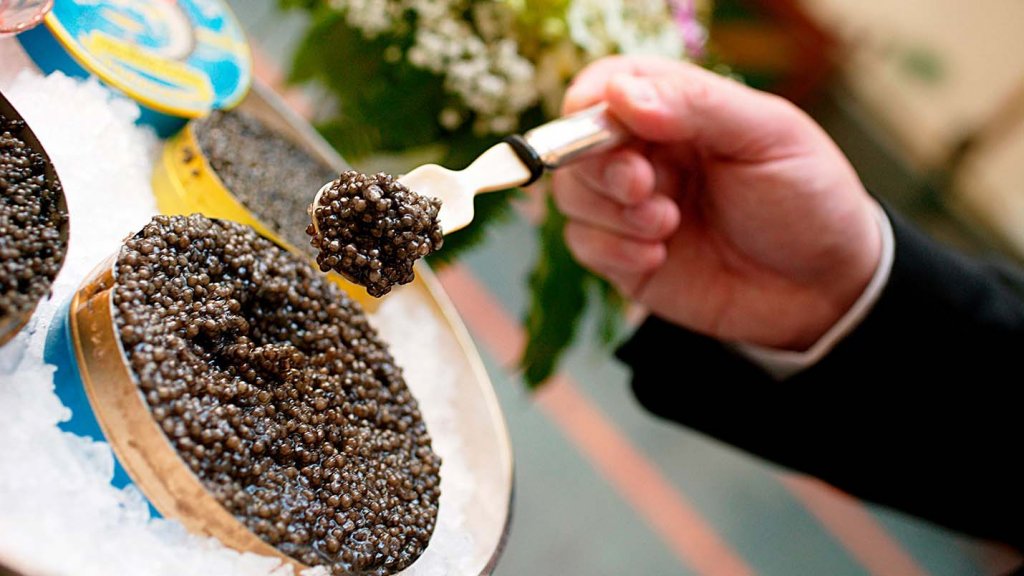}
\imgcell{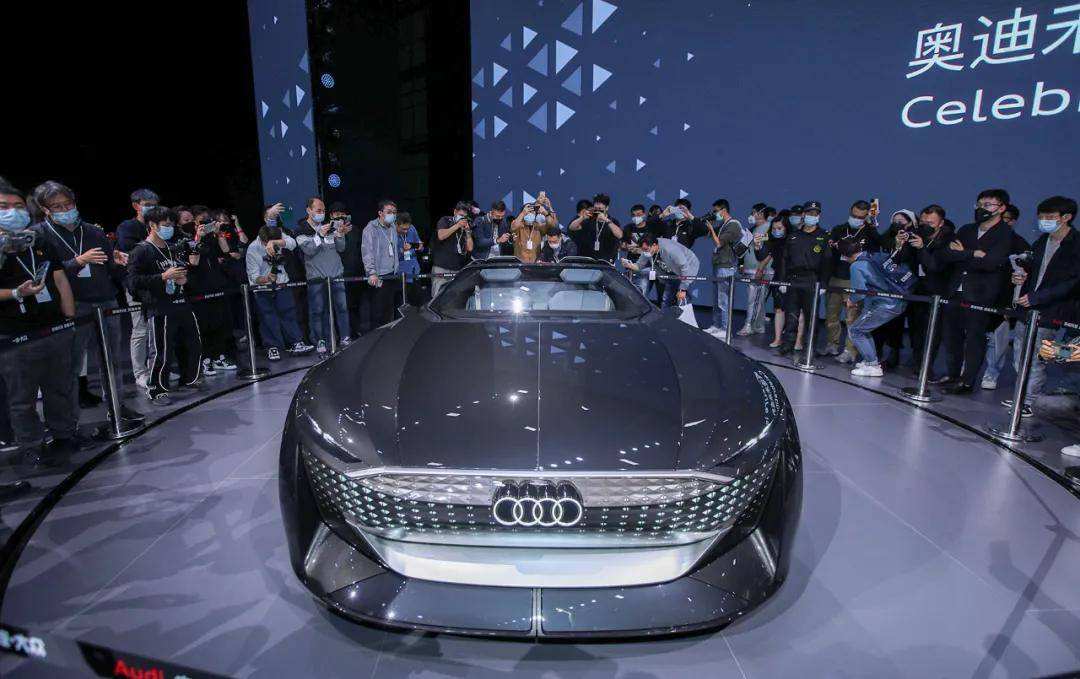}
\imgcell{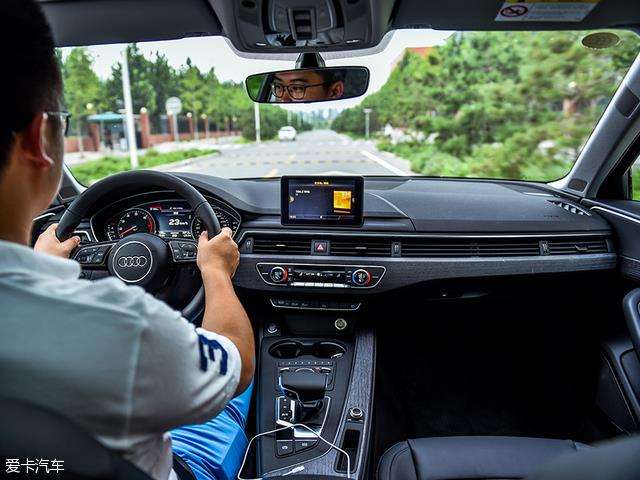}
\imgcell{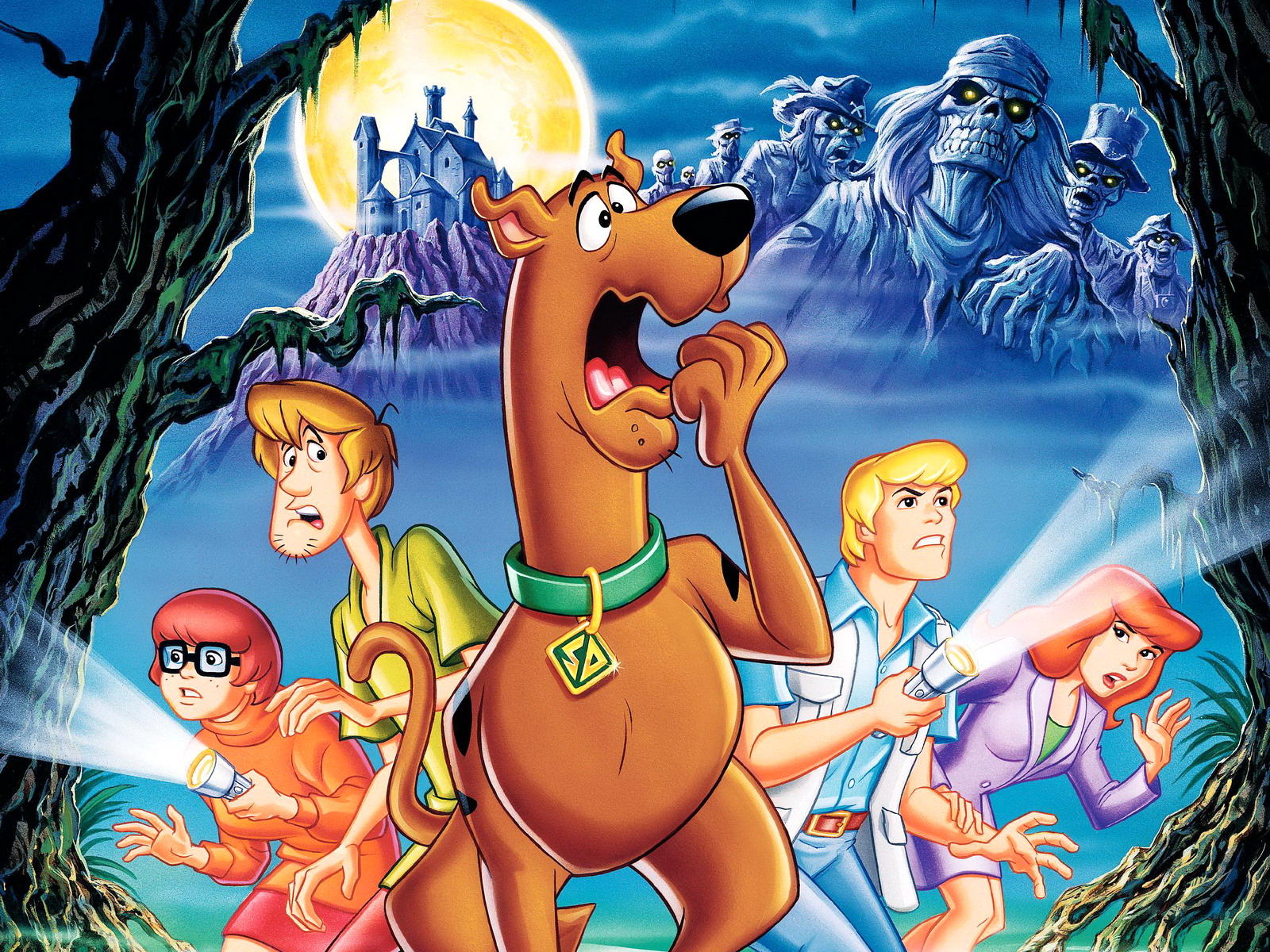} \\[1ex]

% Row 2
\txtcell{\textbf{K-Replay:} A close up of a fast food meal in a box}
\txtcell{\textbf{K-Replay:} A close up of a person cutting up a blackberry}
\txtcell{\textbf{K-Replay:} A concept car is displayed at show}
\txtcell{\textbf{K-Replay:} A man driving a car with a rear view mirror}
\txtcell{\textbf{K-Replay:} A bear in the big blue house} \\[1ex]

% Row 3
\txtcell{\textbf{KRCapVLM:} A \textcolor{darkgreen}{mcdonalds} breakfast sandwich with \textcolor{darkgreen}{scrambled eggs} and \textcolor{darkgreen}{hash browns}}
\txtcell{\textbf{KRCapVLM:} A close up of a person scooping seeds from a \textcolor{darkgreen}{caviar} dish}
\txtcell{\textbf{KRCapVLM:} A \textcolor{darkgreen}{audi} concept car is displayed during the \textcolor{darkgreen}{shanghai} auto show in \textcolor{darkgreen}{shanghai}}
\txtcell{\textbf{KRCapVLM:} A \textcolor{darkgreen}{audi a4} is being driven by a man}
\txtcell{\textbf{KRCapVLM:} A \textcolor{darkgreen}{scooby doo} cartoon with a castle in the background} \\[3ex]

% Row 4
\imgcell{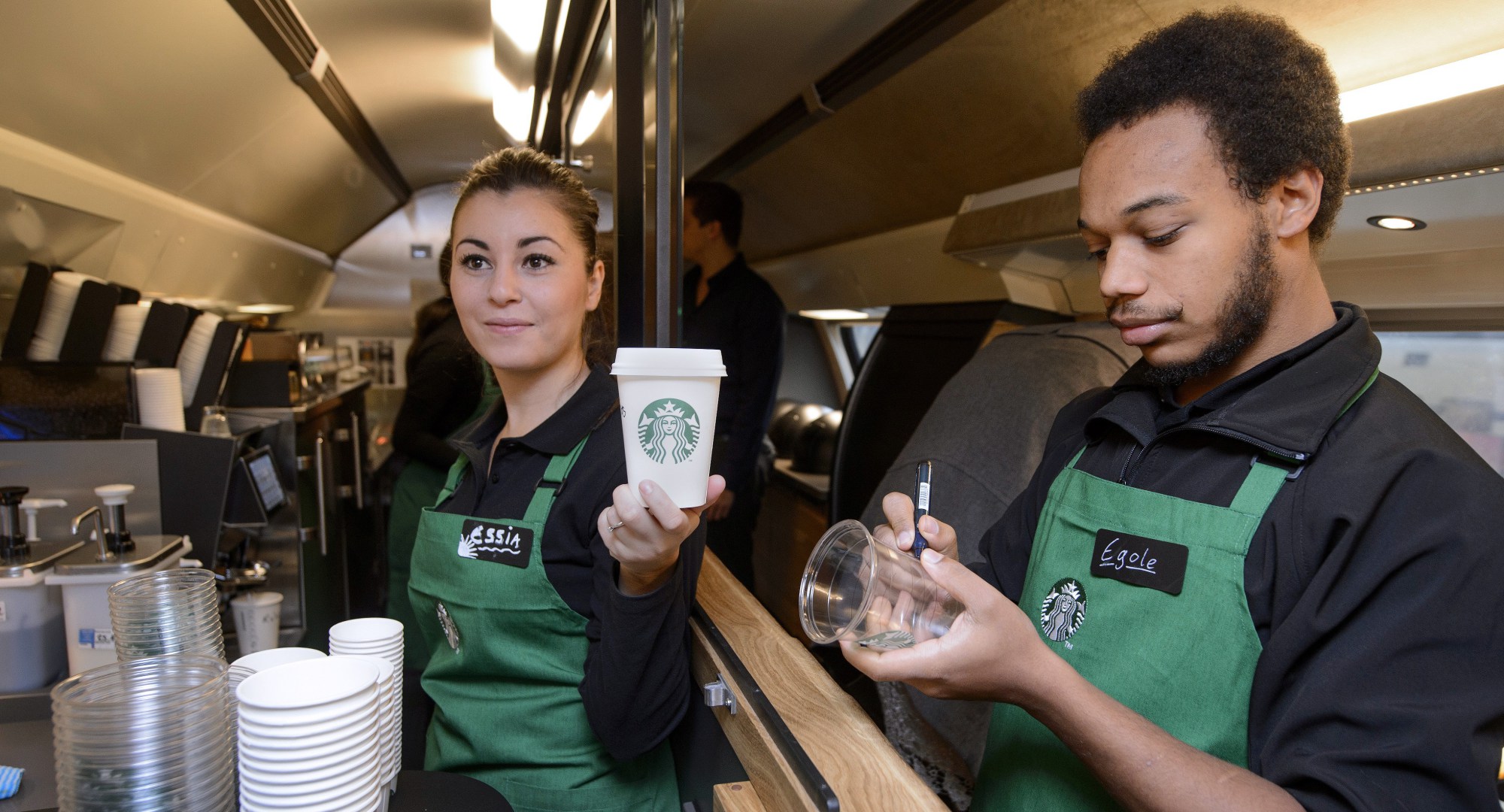}
\imgcell{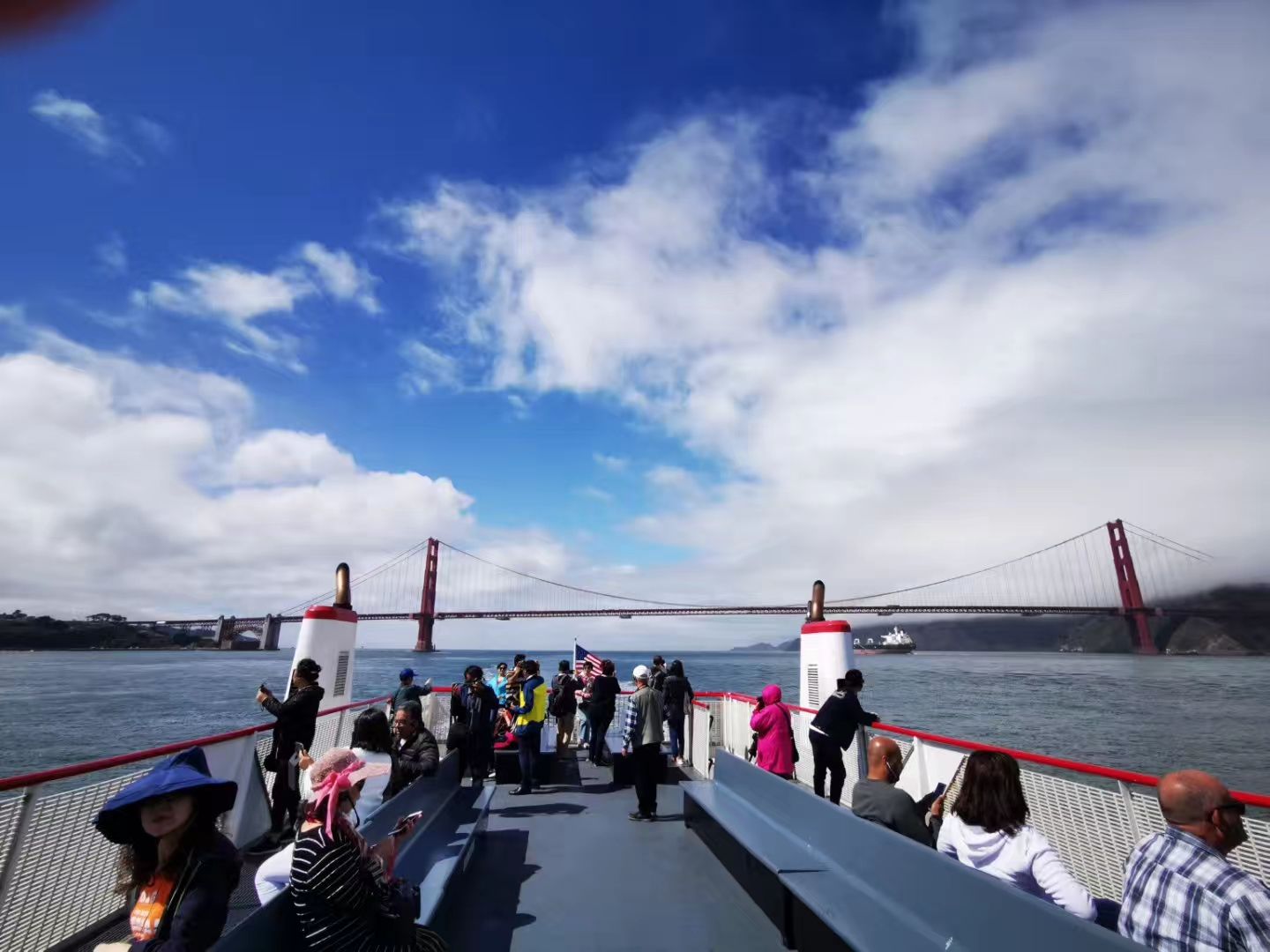}
\imgcell{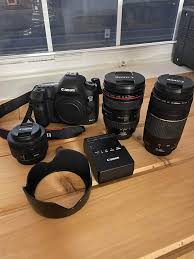}
\imgcell{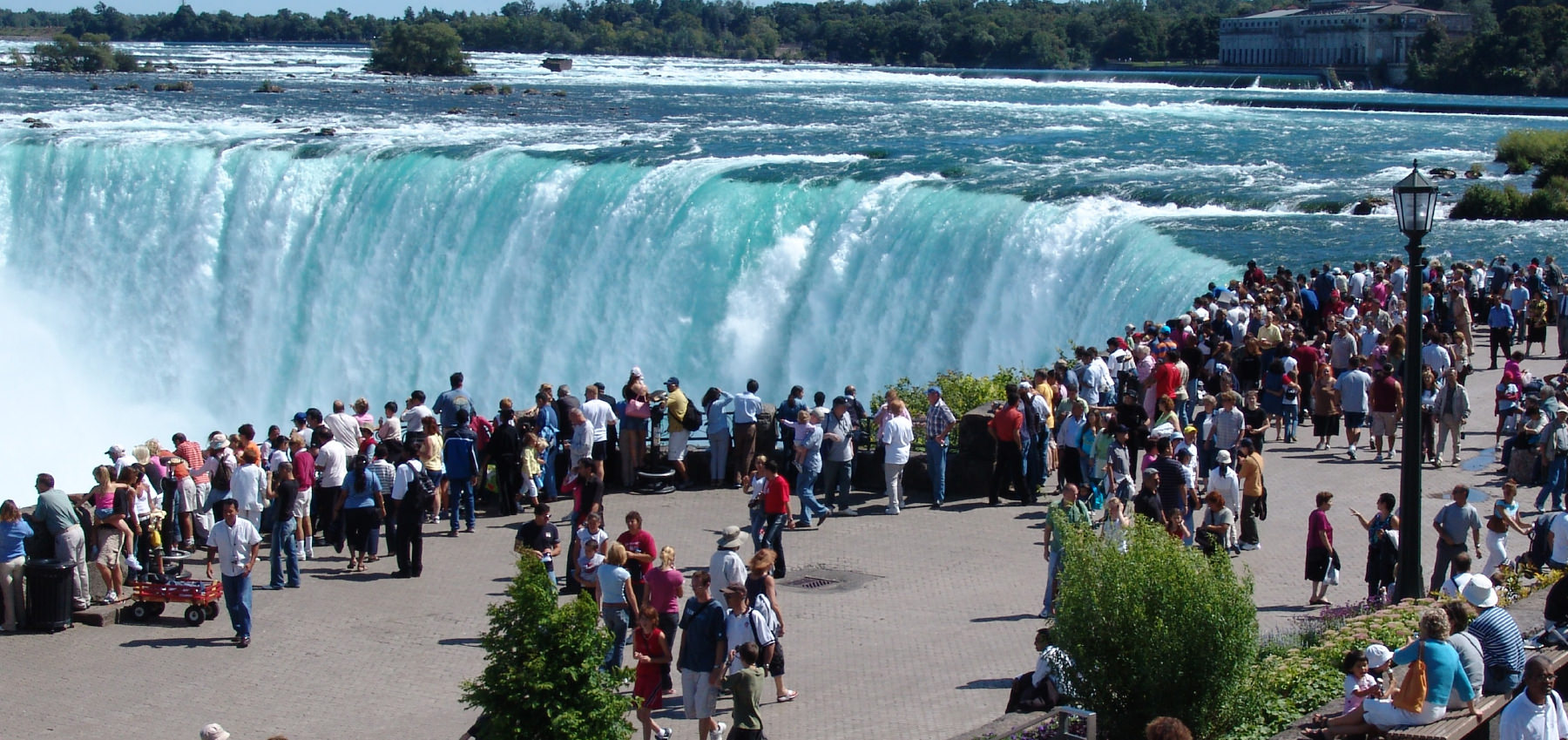}
\imgcell{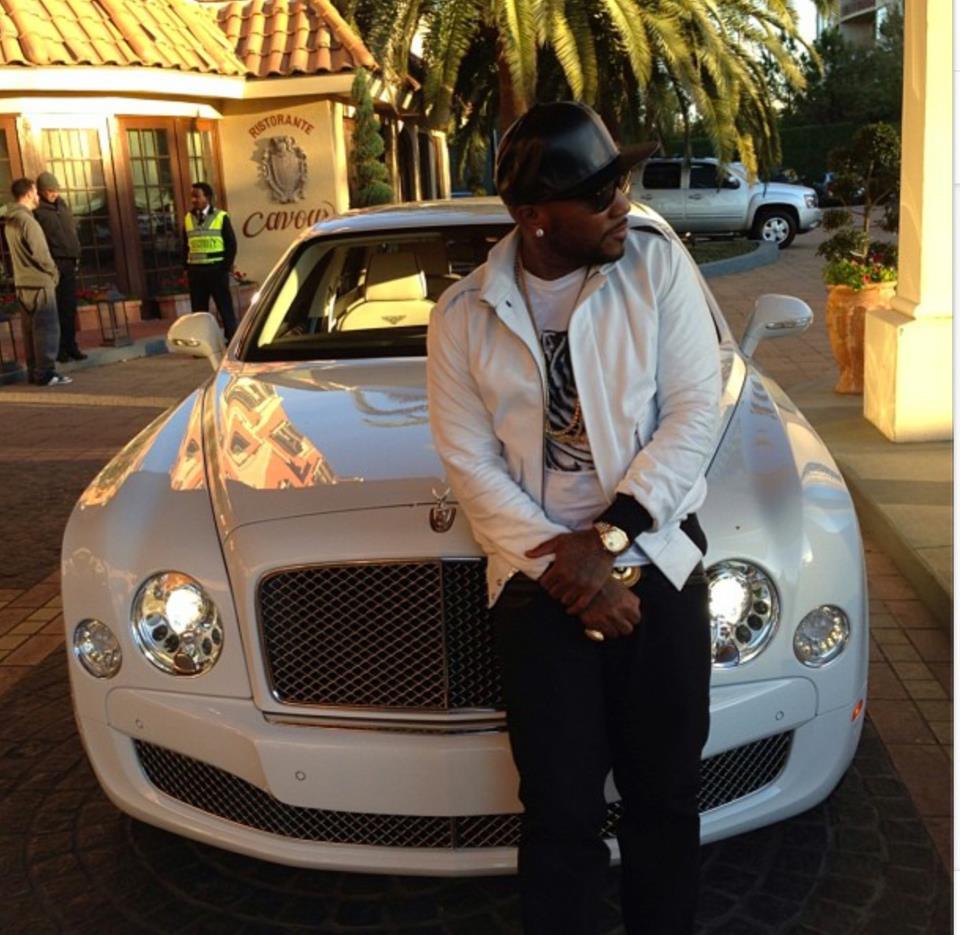} \\[1ex]

% Row 5
\txtcell{\textbf{K-Replay:} A man and a woman wearing a green apron are holding a cup and a glass}
\txtcell{\textbf{K-Replay:} A bridge over the water with a bunch of people on a boat}
\txtcell{\textbf{K-Replay:} A group of cameras sitting on top of a wooden table}
\txtcell{\textbf{K-Replay:} a group of people standing on top of a waterfall}
\txtcell{\textbf{K-Replay:} A man standing in front of a white car} \\[1ex]

% Row 6
\txtcell{\textbf{KRCapVLM:} A man and a woman wearing a \textcolor{darkgreen}{starbucks} apron are holding a cup and a glass}
\txtcell{\textbf{KRCapVLM:} caption: A view of the \textcolor{darkgreen}{golden gate bridge} from a boat}
\txtcell{\textbf{KRCapVLM:} A \textcolor{darkgreen}{canon dslr} and some other cameras on a table}
\txtcell{\textbf{KRCapVLM:} A large group of people standing at the edge of \textcolor{darkgreen}{niagara falls}}
\txtcell{\textbf{KRCapVLM:} A man standing in front of a \textcolor{darkgreen}{bentley car}} \\[3ex]

% Row 7
\imgcell{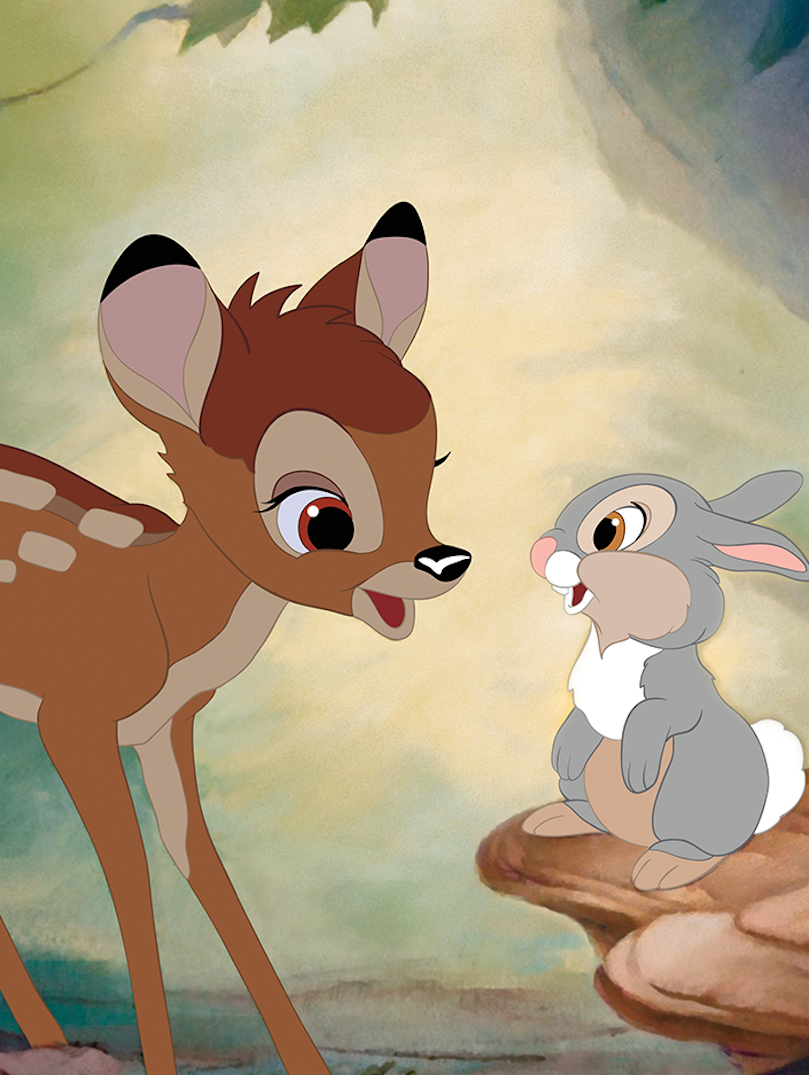}
\imgcell{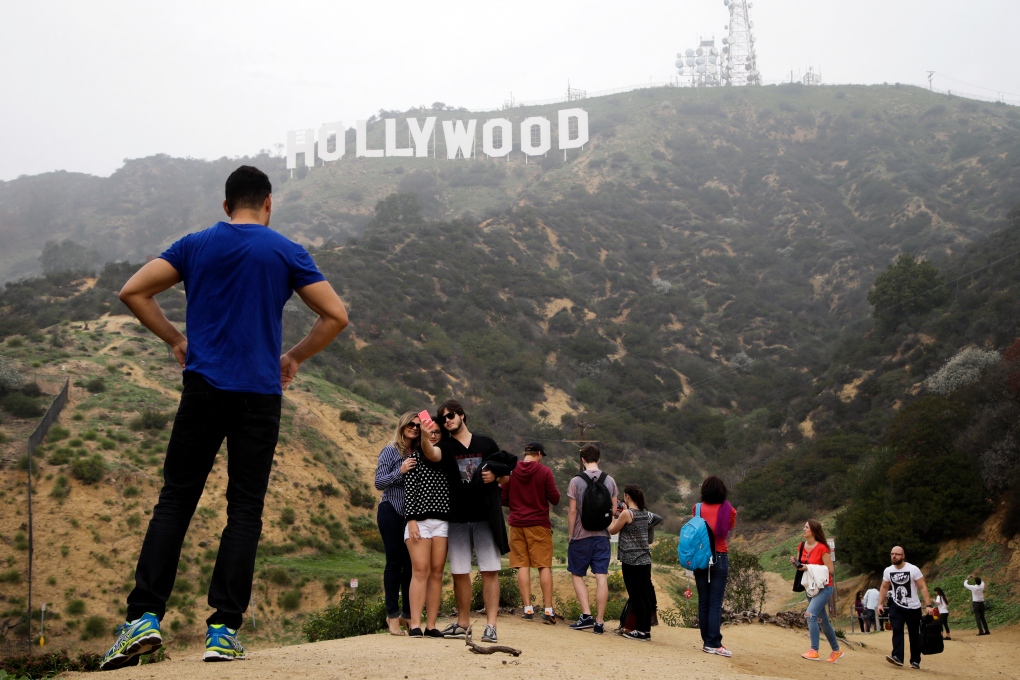}
\imgcell{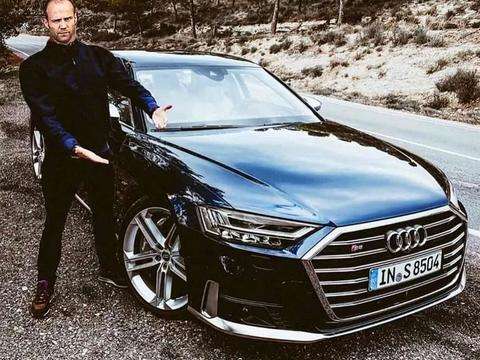}
\imgcell{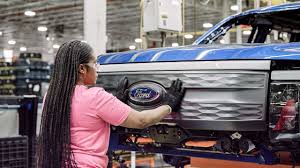}
\imgcell{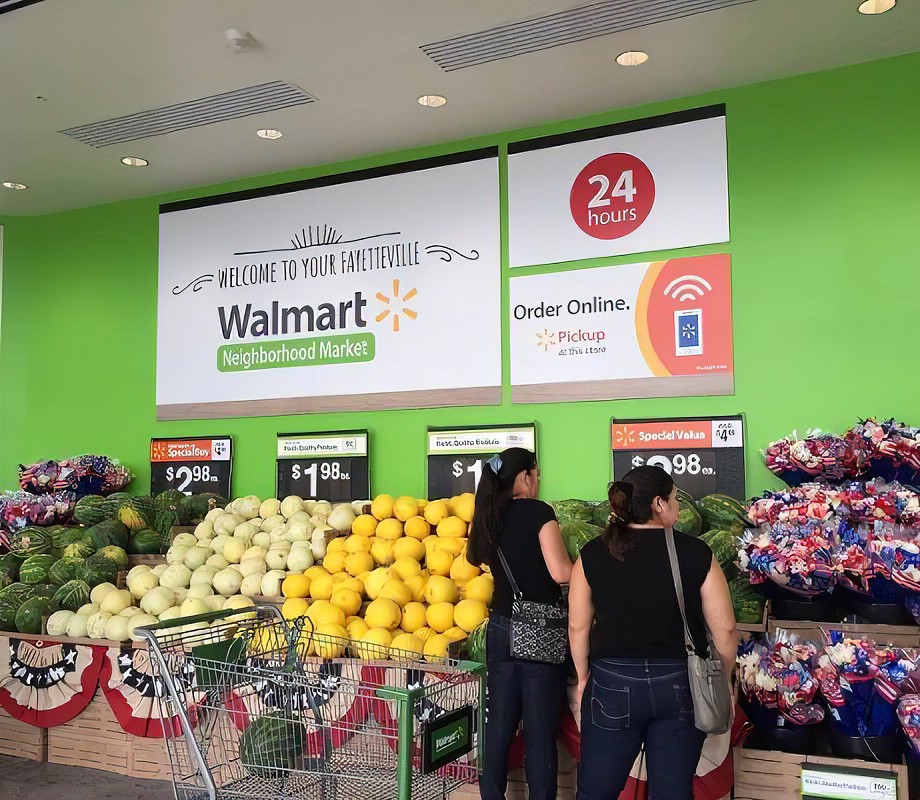} \\[1ex]

% Row 8
\txtcell{\textbf{K-Replay:} A deer and a rabbit looking at each other}
\txtcell{\textbf{K-Replay:} A group of people standing at the top of a hill}
\txtcell{\textbf{K-Replay:} a man standing next to a car on the side of a road}
\txtcell{\textbf{K-Replay:} A woman in a pink shirt is working on a truck}
\txtcell{\textbf{K-Replay:} A group of people standing in front of a store} \\[1ex]

% Row 9
\txtcell{\textbf{KRCapVLM:} A \textcolor{darkgreen}{bambi} deer and a rabbit}
\txtcell{\textbf{KRCapVLM:} A group of people standing at the \textcolor{darkgreen}{hollywood} sign}
\txtcell{\textbf{KRCapVLM:} A man standing next to a \textcolor{darkgreen}{audi} car on a road}
\txtcell{\textbf{KRCapVLM:} A woman in a pink shirt is working on a \textcolor{darkgreen}{ford} truck}
\txtcell{\textbf{KRCapVLM:} two women standing in front of a \textcolor{darkgreen}{walmart} display} \\[3ex]

% Row 10
\imgcell{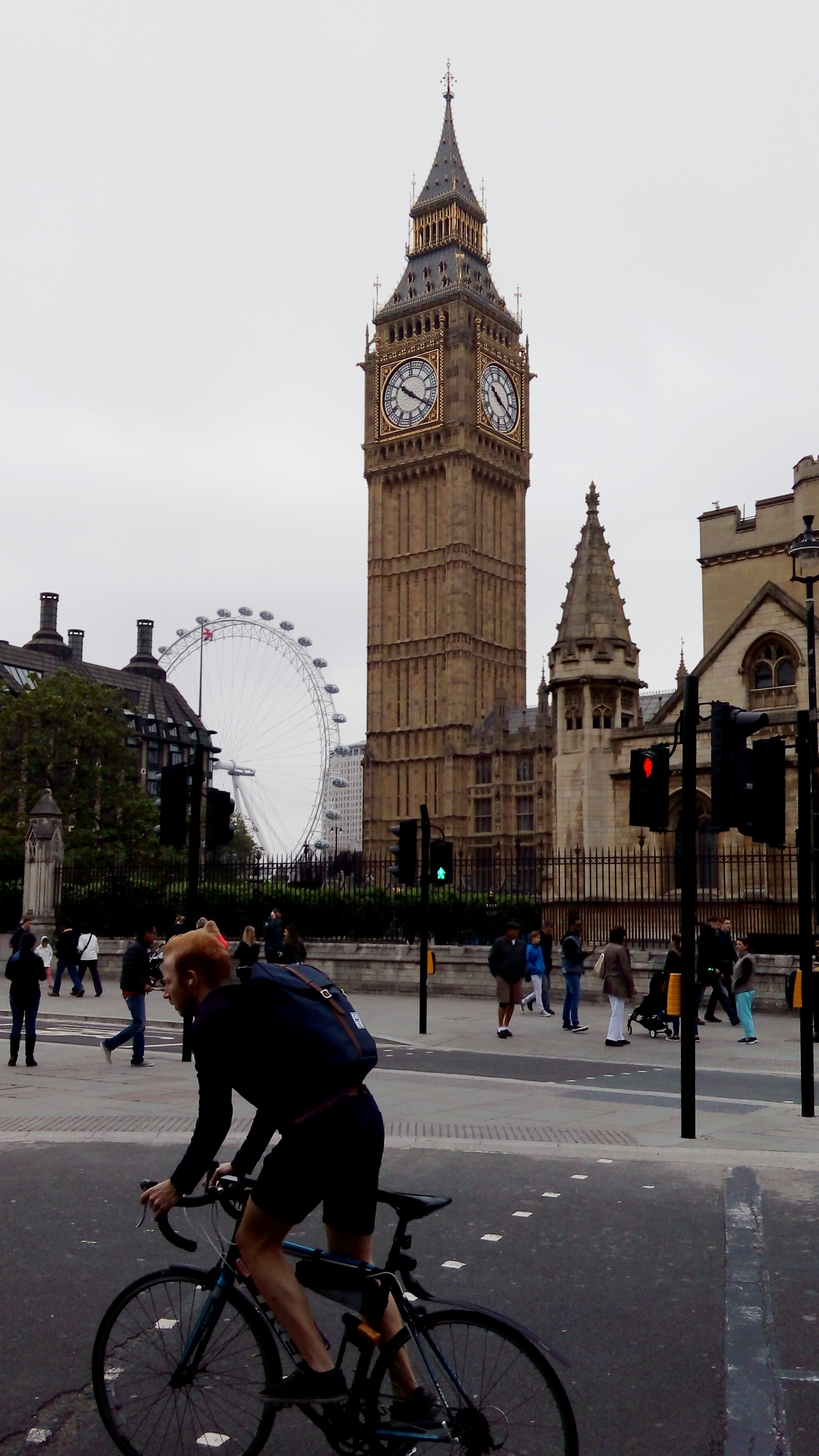}
\imgcell{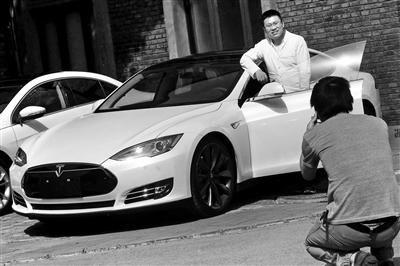}
\imgcell{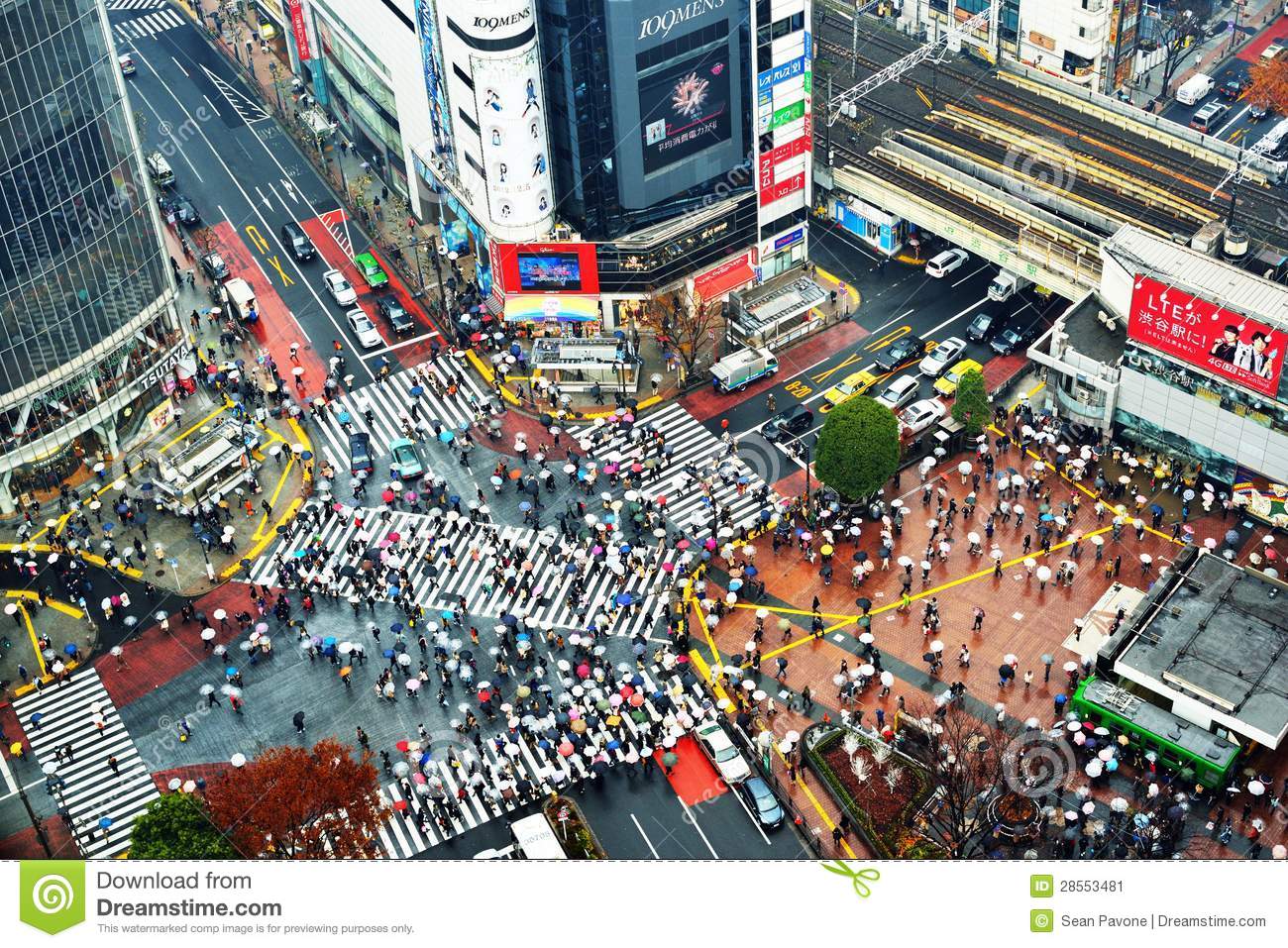}
\imgcell{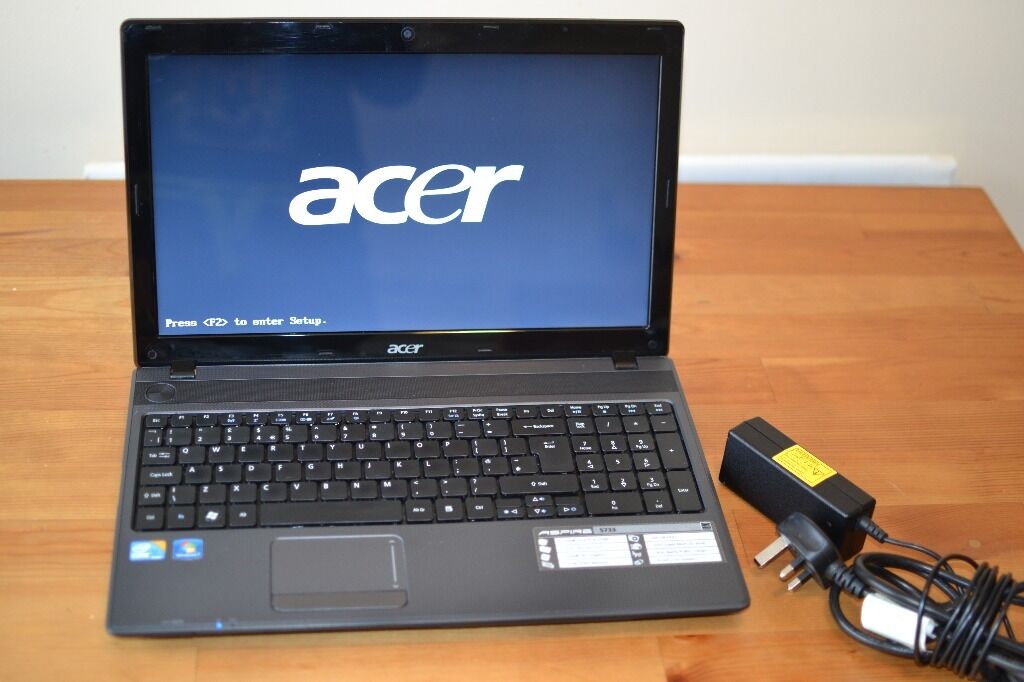}
\imgcell{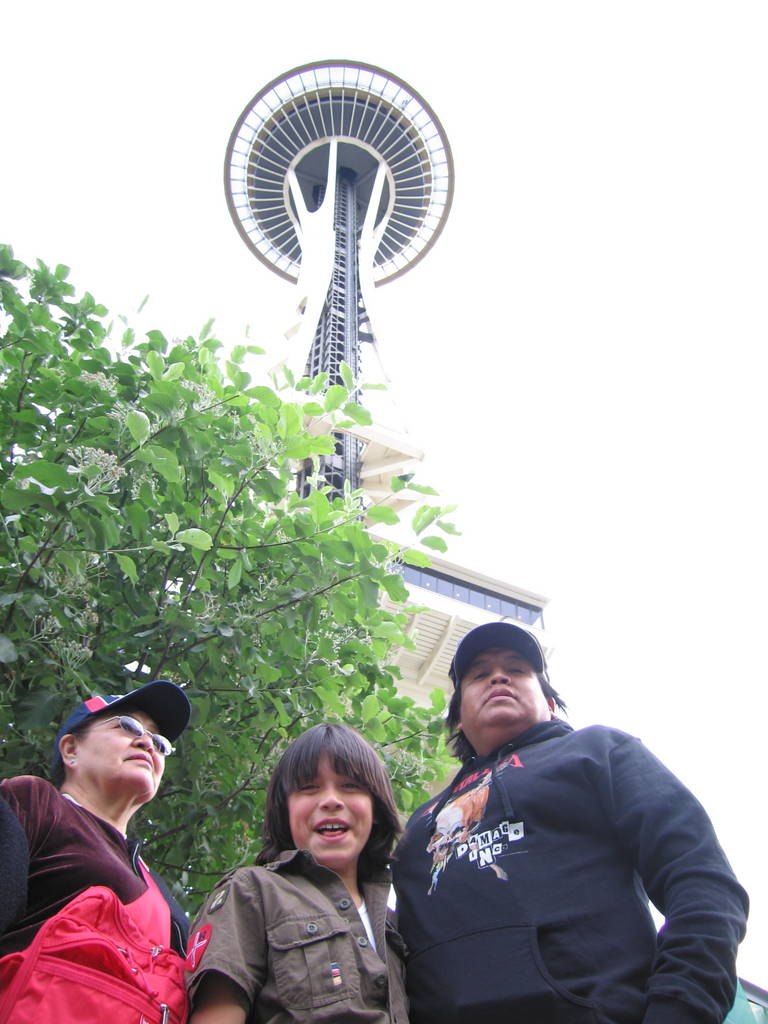} \\[1ex]

% Row 11

\txtcell{\textbf{K-Replay:} A clock tower with a ferris wheel in the background}
\txtcell{\textbf{K-Replay:} A couple of people standing next to a white car}
\txtcell{\textbf{K-Replay:} An aerial view of a shinjuku intersection with a lot of people}
\txtcell{\textbf{K-Replay:} A laptop computer sitting on top of a wooden table}
\txtcell{\textbf{K-Replay:} A group of people standing in front of a tower} \\[1ex]

% Row 12
\txtcell{\textbf{KRCapVLM:} the \textcolor{darkgreen}{big ben} clock tower towering over the city of \textcolor{darkgreen}{london}}
\txtcell{\textbf{KRCapVLM:} A couple of people standing in front of a white \textcolor{darkgreen}{tesla}}
\txtcell{\textbf{KRCapVLM:} A \textcolor{darkgreen}{shibuya scramble crossing} in \textcolor{darkgreen}{tokyo} city}
\txtcell{\textbf{KRCapVLM:} A \textcolor{darkgreen}{acer} laptop computer sitting on top of a wooden table}
\txtcell{\textbf{KRCapVLM:} A group of people standing in front of the \textcolor{darkgreen}{space needle}} \\[3ex]

\caption{Comparison of captions generated by K-Replay \cite{cheng2023beyond} and KRCapVLM (Ours) across diverse image examples.}
\label{fig:caption_comparison}
\end{figure*}

In Figure \ref{fig:caption_comparison}, we showcase the improvements made by KRCapVLM model over the original K-Replay captions. Each image is presented with the captions generated by the model, both before and after the enhancement. The KRCapVLM model significantly improves upon the original K-Replay captions by offering more detailed descriptions, incorporating real-world concepts, which are highlighted in green for emphasis.

\subsection{Cross-Dataset Evaluation and Generalization Results} 
In addition to the improvements on the KnowCap dataset,  we assess the model on unseen KnowCap categories, a portion of the test set containing 120 categories and 520 images that were not present in the replay data. This enables us to evaluate the model’s capacity to adapt to new, unseen scenarios. The results are illustrated in Table~\ref{table:unseen_comparison}. Starting with the OFA zero-shot baseline, we observe a CIDEr score of 36.9 and the recognition accuracy of 39.2\%, which serves as the starting point for our evaluation. After fine-tuning with OFA-Finetuned, CIDEr increases to 42.0, and the recognition accuracy remains stable at 39.0\%, confirming the benefits of fine-tuning on this dataset.

When K-Replay is added to the fine-tuned model, we see a remarkable jump in CIDEr to 83.9, and a noticeable increase in Recognition Accuracy to 45.8\%. This result demonstrates that K-Replay is not only helping the model retain learned knowledge but also significantly improving its performance on the unseen categories. We also demonstrate that introducing the learning rate scheduler (+Scheduler) boosts the CIDEr score further to 91.0 and the recognition accuracy to 55.6\%, showing that adjusting the learning rate aids in better fine-tuning and model convergence.

However, after adding beam search (+Beam), we see a slight drop in the CIDEr score, which decreases to 82.6, but the recognition accuracy increases to 57.7\%. This suggests that although beam search improves generation quality, it helps the the recognition accuracy even further, leading to higher recognition accuracy. Finally, adding attention layers (+Attention) leads to a slight improvement in the CIDEr score, bringing it to 84.4, but the recognition accuracy decreases slightly to 55.0\%. The attention layers allow the model to focus on more relevant visual features, which enhances the overall concept grounding, though the recognition accuracy slightly drops compared to the beam search configuration.

\subsection{Comparison with State-of-the-Art Methods} 
Catastrophic forgetting remains a major challenge in the fine-tuning of VLMs, as the process of adapting to downstream tasks like image captioning can result in the loss of pre-trained knowledge. In this section, we compare our proposed methods with several previous approaches aimed at mitigating catastrophic forgetting, particularly in the context of concept-aware image captioning. The first one is EWC \cite{kirkpatrick2017overcoming} which introduces a regularization term that penalizes significant changes to parameters important for the original task, aiming to preserve prior knowledge during adaptation. The second one is Recall and Learn \cite{chen2020recall} that employs regularization weights to help the model retain previously learned knowledge. The third one is Child-Tuning \cite{xu2021raise} which focuses on fine-tuning only a specific set of parameters that are considered essential for the downstream task. This approach helps enhance the model's generalization and minimizes the risk of forgetting. The fourth is Adapter \cite{gao2021clip} which uses lightweight bottleneck modules to inject task-specific parameters while leaving the original pre-trained model largely intact, thus preventing overfitting and catastrophic forgetting.

In Table~\ref{table:comparison}, we provide quantitative comparisons of these methods on the KnowCap dataset. While all methods introduce strategies to alleviate forgetting, their effectiveness varies considerably. Notably, even strong baselines like EWC and Child-Tuning struggle to maintain both high recognition accuracy  and CIDEr scores simultaneously. By contrast, K-Replay shows substantial improvements, especially in recognition-focused metrics. Our proposed KRCapVLM further pushes these gains, achieving the highest recognition accuracy and maintaining competitive scores across all metrics. This highlights the robustness of our approach in preserving concept-awareness during fine-tuning, outperforming previous techniques without compromising language quality.

\begin{table}[t!]
\centering
\caption{Comparison of different state-of-the-art methods for catastrophic forgetting on the KnowCap dataset.}
\label{table:comparison}
\setlength{\tabcolsep}{6pt} % Adjust column spacing as needed
\begin{tabular}{lcccccc}
\toprule
\textbf{Model} & \textbf{B1} & \textbf{B4} & \textbf{M} & \textbf{R} & \textbf{C} & \textbf{Rec} \\
\midrule
EWC~\cite{kirkpatrick2017overcoming} & 56.9 & 21.8 & 19.1 & 42.0 & 73.6 & 30.40\% \\
Recall and Learn~\cite{chen2020recall} & 53.7 & 20.1 & 18.1 & 39.3 & 70.6 & 37.20\% \\
Child-Tuning~\cite{xu2021raise} & 55.8 & 21.7 & 18.8 & 41.5 & 74.7 & 33.80\% \\
Adapter~\cite{gao2021clip} & 54.4 & 20.5 & \textcolor{red}{17.6} & 40.1 & \textcolor{red}{63.7} & \textcolor{red}{30.40\%} \\
Vanilla Fine-tuning~\cite{cheng2023beyond} & \textcolor{red}{35.6} & \textcolor{red}{10.8} & 12.1 & \textcolor{red}{26.6} & 41.7 & 38.50\% \\
\midrule
K-Replay & \textcolor{blue}{58.8} & 22.7 & 20.2 & 43.0 & 90.3 & 50.40\% \\
+Scheduler & 58.6 & \textcolor{blue}{23.9} & 20.9 & \textcolor{blue}{44.1} & \textcolor{blue}{92.6} & 54.20\% \\
+Beam & 57.7 & 21.9 & 20.4 & 42.5 & \textcolor{blue}{92.6} & \textcolor{blue}{63.30\%} \\
+Attention & 58.1 & 22.1 & \textcolor{blue}{20.5} & 43.2 & 92.0 & 58.90\% \\
\bottomrule
\end{tabular}
\end{table}

\section{Conclusion} 
In this work, we proposed KRCapVLM that aims to improve both the caption generation quality and knowledge recognition. The proposed approach replaces greedy decoding with beam search, integrates attention-based modules into the image encoder, and employs training schedulers to enhance stability and promote smoother convergence. The proposed contributions resulted in significant performance improvements. The proposed KRCapVLM model was evaluated on both the COCO and KnowCap datasets, showing a substantial increase in recognition accuracy, while maintaining or slightly improving the overall captioning performance, as reflected by the CIDEr scores. More importantly, the model demonstrated strong generalization capabilities on unseen knowledge categories, with recognition accuracy improving from a baseline of 45.80\% to 57.69\%. These results underscore the effectiveness of KRCapVLM in enhancing both the accuracy and robustness of knowledge-aware image captioning.

\bibliographystyle{elsarticle-num}
\bibliography{library-fixed}

\end{document}